# A RAPID WILDFIRE RESPONSE SYSTEM BASED ON DRONES

## U LENGHOU, BAI YUQIAN AND LUO QIFAN

ABSTRACT. The wildfires in Australia make people suffer the losses caused by the devastating disasters. To protect people's security and the country's development in vulnerable situations, we are expected to establish a set of efficient and sustainable fire-rescuing response mechanisms and tackle the problems faced, including optimization of loadings and quantity of drones, time adaptability of the model, and the locations of radio-repeater drones. We determined the location of mobile emergency operation centers (EOC), which conduct drone launches and deploy rescue personnel. After figuring out where the EOCs are located, we can establish the Drones Planning-programming Budgeting   Model to detect the configuration of drones, including optimal numbers and mix of SSA drones and Radio Repeater drones. To estimate the adaptability of our model in the next ten decades, we build a Holt-Winters seasonal model. We utilize the Drones Deployment Optimization model to optimize the locations of hovering VHF/UHF radio-repeater drones. In addition,   sensitivity analysis of the model is tested.

## 1 INTRODUCTION

The ongoing outbreak of bushfire in Australia is one of the most long-lasting, wide-spreading, and far-reaching catastrophic fires in the world. Exacerbated by climate change, wildfires broke out during a severe drought and persistent heatwave. The devastating wildfires in every state is a world concerning environmental issue, which needs to be solved by a reasonable operation.

Rapid wildfire response is carried out with the assist of the sensor and communication system, including drones for surveillance and situational awareness (SSA), handheld two-way radios for a status report, and repeaters for radio range extension.

As the elementary equipment for fire rescue, SSA drones are able to carry high definition & thermal imaging cameras and telemetry sensors to transmit data from front line wearable devices to the Emergency Operations Center (EOC). The deployed personnel use it to perceive and understand the current situation of the fire, predict the development of fire in a short time and systematically analyze, collect and share the data for the sake of effective guidance for "boots on the ground" towards teams.

A RAPID WILDFIRE RESPONSE SYSTEM BASED ON DRONES

Handheld radio communication equipment (HRCE), working in VHF / UHF bands, can achieve two-way communication between firefighters and control commanders. Determined by the distance and terrain, the maximum power of the handheld radio transmitter is 5 watts, which enables a nominal range of 5 km on the barrier-free regions and 2 km in the urban area containing high buildings and large mansions, with little significant effect from the weather. The weather has little effect on it.

A repeater is an unattended transceiver, whose function is to automatically rebroadcast the signal received in a specific range, thus dramatically expands the range of low-power radio. A hovering UAV carries a 10-watt repeater, with the weight of 1.3kg, can expand the range to 20km. Distance and physical topography are also the significant influence factors of repeater range. Even if it is limited by various obstacles, its range is still greatly larger than that of low-power handheld radio without a repeater.

Recently, Akme Corporation's prototype WileE–15.2X hybrid drone is projected to cost approximately $10,000 (AUD). The new product has a flight distance of 30km, a maximum flight speed of 20 m/s, a maximum flight time of 2.5 hours and a charging time of 1.75 hours. The carrying devices of these new hybrid drones can be selected from a radio repeater or video & telemetry capability to cooperate with the firefighters' handheld radio for the best informative transmission.

The linkage of these scientific and technological products enables EOC to monitor the real-time variation of fire, so as to make overall arrangements for fire rescue and maximize fire-fighting efficiency.

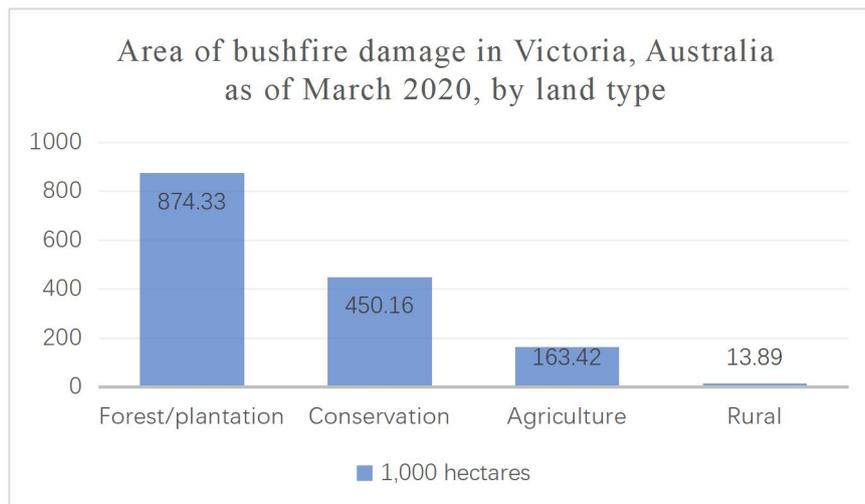

**Fig 1.1** Area of bushfire damage in Victoria, Australia as of March 2020, by land type

A RAPID WILDFIRE RESPONSE SYSTEM BASED ON DRONES

According to figure1.1, we are told that there are 874.33 thousand hectares of forest and plantation are damaged by bushfire while 450.16 thousand hectares of the conservation area are devastated. Plus, there are 163.42 thousand hectares of agriculture area and 13.89 thousand hectares of rural area are damaged by bushfire. Therefore, bushfires cause serious property damage.

Even worst, in February 2020, at least 33 people have been killed, 3094 houses have been destroyed, 17 million hectares of land have been burned down, and one billion animals have died in the 2019-2020 bushfire [1]. The wildfire in Australia has threatened the security of people's property and the life which is an urgent need to solve this major environmental crisis. Therefore, the importance and emergency of a feasible strategy to Australia's bushfires cannot be underestimated.

In previous work, a reliable, resilient, and quickly deployable emergency communication network was to construct a drone-assisted emergency Wi-Fi network is proposed to expedite the rescue operations by guiding the survivors to the nearest rescue camp location [2]. The three-dimensional modeling and orthophoto mapping have the potential to facilitate disaster managers with more precise damage assessment [3]. However, the geographical environment of Australia is complex. These rescue methods may not be completely suitable for Australia, which has certain limitations. It is urgent to develop a rescue plan that can be applied to this special situation.

Our team aims to build a mathematical model to formulate an effective strategy for purchasing drones to the Country Fire Authority (CFA), and to illustrate how it can adapt to the fire situation in the next decade. Then we are supposed to determine the optimal hovering location of the drones. Ultimately, a business letter is provided to the Victorian government to elaborate on the reasons for the budget application. Through the above measures, we can minimize the negative impact of this disaster on Australia.

## 2 PREPARATION

To preferably solve the three problems in the requirement, in the establishment of our "rapid bushfire response" mechanism, we intend to find a mobile EOC location for each fire site, which can be used as the front-line command center for an emergency response to conduct drones launch and deploy rescue personnel. This section describes how we use historical data to select mobile EOCs for different fire locations.

According to the numerical data collected by satellite for fire disasters during 2019-2020, we assumed that small fires do not need to deploy EOC, so we designed to exclude the fires whose fire radiative power is lower than 90 watts. Then, we draw figure 2.1. Fire sites are marked on the map so that we can discuss the selection of mobile EOC locations based on this picture.

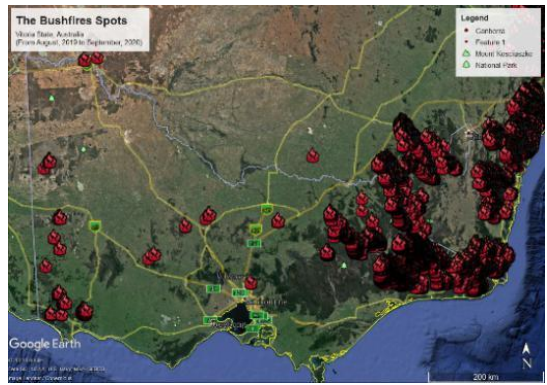

**Fig 2.1** The bushfires spots in Victoria

The flame temperature of forest fire is generally 800-1000 ℃, which is significantly higher than that of ordinary fire. Due to its wide range of conflagration, it takes more time to control the fire. The location that can be used as a mobile EOC needs to have a complete power supply system to ensure the normal operation of communication, monitoring, analysis, display, and other equipment. On the other hand, the city where the mobile EOC is located needs to have a competitive logistics supply capacity to provide support for emergency response personnel for the reason that severe fire necessitates a greater quantity of firefighters and facilities. Therefore, we assume that only cities with more than 10000 (modifiable) populations can be used as mobile EOC for forest fire emergency response.

After searching for data from the Australian Bureau of statistics, we marked the cities with a population of more than 10000 in Victoria Province of Australia on the map and drew figure 2.2

A RAPID WILDFIRE RESPONSE SYSTEM BASED ON DRONES

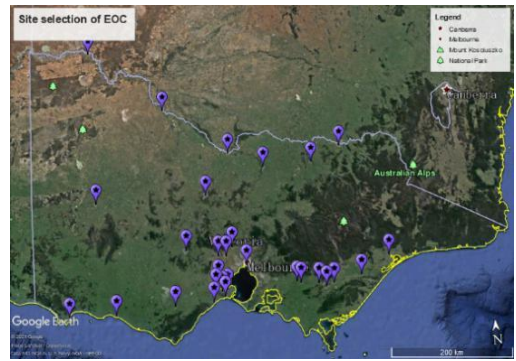

**Fig 2.2** The site selection of EOC candidate in Victoria

With the aim of the lowest cost and the fastest efficiency, the location of mobile EOC must be the closest to the fire location among all the alternative locations, which can reduce the number of drones carrying repeaters, so as to improve the communication efficiency and reduce the cost. We assume that the terrain obstruction is negligible. Launched in the mobile EOC, the drones usually move in a straight line to ensure that they can rescue the fire with the shortest distance and the fastest speed.

Next, we analyze the different fire situations presented by the historical data: As for sparsely distributed fire sites, we regard the sites of the fire and the mobile EOC as points of the plane rectangular coordinate system. Because the fire area of discrete fire regions is small, the distance between the fire region and the location of mobile EOC is much larger than the area of the fire-affected region and EOC. When calculating the distance between them, the area of them has little influence on the calculation results, so they are negligible. To simplify the model, we assume that they are all points in the coordinate system.

As for densely distributed fire sites, we assume that they occurred at the same time. We select several points in the area to calculate the average distance between fire points and the mobile EOC candidates, which will be computed as the distance between the mobile EOC and the fire site. According to the map analysis, there are dense trees near the densely distributed fire sites. Consequently, influence is felt in a wider range. Because the error caused by taking the average value as the distance is far less than the distance between the mobile EOC and the fire site. For the fire area with a large burning area, the distance between the fire location and EOC can not be accurately calculated as the discrete points. In order to simplify the model, we use the average distance of dense points to express the distance between the fire location and EOC.

Finally, we draw figure 2.3 with the aid of the measurement tool of Google Earth.

A RAPID WILDFIRE RESPONSE SYSTEM BASED ON DRONES

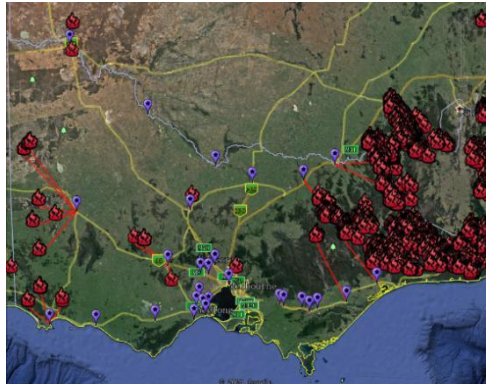

**Fig 2.3** The bushfires spots and the EOC allocation

According to the calculation on Google Earth, we can figure out the cities that can be used as mobile EOC. They are Mildura, Horsham, Portland, Ballarat, Melbourne, Wangaratta, Wodonga, Sale, and Bairnsdale. These cities and their longitude and latitude coordinates are shown in figure 2.4.

| latitude | longitude | Name of the city |
|----------|-----------|------------------|
| -34.206 | 142.136 | Mildura |
| -36.716 | 142.199 | Horsham |
| -38.346 | 141.604 | Portland |
| -37.549 | 143.850 | Ballarat |
| -37.840 | 144.946 | Melbourne |
| -36.358 | 146.312 | Wangaratta |
| -36.124 | 146.876 | Wodonga |
| -38.099 | 147.066 | Sale |
| -37.833 | 147.616 | Bairnsdale |

**Fig 2.4** List of the site selection of EOCs and their longitude and latitude

# 3 OUR MODEL

In this section, we will describe our model for the three problems mentioned in the introduction. The first model is responsible to determine the quantity of SSA drones and repeater drones. Our second model is responsible to determine what equipment cost will increase if fires become severe in the next decade. The third model is in charge of the deployment of drones.

## 3.1 Drones Planning-programming Budgeting Model

### 3.1.1 Assumption

In order to simplify the model, we take a square with an area of $10\sqrt{2} \times 10\sqrt{2}$ square kilometers as the minimum unit to calculate the fire spread area to calculate the number of drones with repeaters;

We assume that the range of drones with repeaters is invariably 20 km, as shown in the figure, there are no significant obstacles like lofty mountains, which hinder the transmission of the signal. When the drones flys high enough, the path of transmitting is mostly in open airspace.

Given that we are the emergency team dealing with forest fires, to simplify the model, we assume that trees are the only obstacles to the signal transmission of handheld radios and front-line wearable devices.

### 3.1.2 Notations

| Symbol | Definition |
|--------|------------|
| $\rho$ | The density of trees in the forest, number of trees per unit area |
| $k$ | Scale factor |
| $d_0$ | Fire affected area |

A RAPID WILDFIRE RESPONSE SYSTEM BASED ON DRONES

| | |
|---|---|
| $d$ | Number of drones carrying repeaters above the fire site |
| $S$ | The number of drones carrying repeaters on the connecting line between the fire site and mobile EOC |
| $q_1$ | Number of drones carrying repeaters for communication |
| $q_2$ | The straight-line distance between fire site and mobile EOC |
| $l$ | The height of the hovering drones |
| $Q_1$ | The maximum distance of signal transmission from handheld radio and front-line wearable devices on the flat and barrier-free ground |
| $h$ | The maximum distance of signal transmission of handheld radio and front-line wearable equipment on specific terrain |
| $r$ | Range of data received by the drones carrying video and telemetry capability |
| $m$ | Number of squares of the drones carrying video and telemetry capability |
| $a_i$ | The side length of the i$^{th}$ square, $i \in 1, 2, ..., m$ |
| $t$ | The coverage area of a drone carrying video and telemetry capability |
| $s_0$ | Number of drones carrying video and telemetry capability for monitoring |
| $Q_2$ | Operation time of drones carrying video and telemetry capability for monitoring |
| $n$ | The total number of forest fires in Victoria State in the past year |
| $A$ | The total fire-affected area in Victoria State in the past year |
| $\overline{A}$ | Average fire-affected area, $\overline{A} = \dfrac{A}{n}$ |
| $T$ | The total duration of forest fires in Victoria State in the past year |
| $\overline{T}$ | The average duration of the forest fire, $\overline{T} = \dfrac{T}{n}$ |
| $\overline{l}$ | The average distance between fire site and mobile EOC in the past year |
| $l_j$ | Distance between j$^{\text{th}}$ fire site and mobile EOC in the past year, $j \in 1, 2, ..., n$ |
| $\overline{q_1}$ | The average number of drones with repeaters used in response to each forest fire in the past year |
| $\overline{q_2}$ | The average number of drones with video and telemetry capability used in response to each forest fire in the past year |
| $Q_3$ | The number of drones with repeaters that need to be purchased in response to next year's bushfire |
| $Q_4$ | The number of drones with video and telemetry capability that need to be purchased in response to next year's bushfire |
| $C$ | In order to cope with the forest fire next year, the ratio of the number of drones with repeaters to the number of drones with video and telemetry capability needs to be purchased |
| $rec_{i,j}$ | Rectangles in figure 3.1.4.2.1. |

## 3.1.3 Model construction

### 3.1.3.1 The influence of topography on the range of communication equipment

When a radio signal propagates in the forest, the more obstacles it encounters on the way, the closer the range of the signal. Therefore, we can establish the following mathematical model:

$$d = k\left(1-\rho\right)d_0$$

(3.1.3.1.1)

## 3.1.4   Relationship between the number of drones and fire size

### 3.1.4.1 Number of drones carrying repeaters for communication

When dealing with a forest fire, drones with repeaters are mainly distributed over the fire site and on the connecting line between mobile EOC and fire site.

In order to have enough repeater drones flying above the fire sense, we first separate the fire sense into rectangles whose area is $10\sqrt{2} \times 10\sqrt{2}$ km². It is the maximum area that a drone can fly. Yet, it is illustrated in figure 3.1.4.2.1.

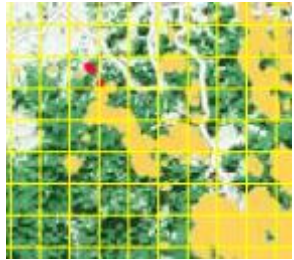

**Fig 3.1.4.2.1** Rectangle above fire sense

If the rectangle is covered by yellow, then we indicate this rectangle $rec_{i,j}$ is equal to 1. Else if it is not covered by yellow, we indicate this rectangle $rec_{i,j}$ is equal to 0. While $rec_{i,j}$ is 1, it means it is the position of drones. If $rec_{i,j}$ is equal to 0, it means it is not necessary to deploy drones above it. Then we can obtain formula 3.1.4.1.1.

$$q_1 = \sum_{i=1}^{n}\sum_{j=1}^{n}(rec_{i,j})$$

(3.1.4.1.1)

### 3.1.4.2 Number of drones carrying video and telemetry capability for monitoring

According to the Pythagorean theorem, we obtain the relationship between the range of data received by the drones carrying video & telemetry capability and the hovering height of the drones, and the signal transmitting distance of the front-line personnel's wearable device in a specific terrain:

$$r = \sqrt{d^2 - h^2}$$

(3.1.4.2.1)

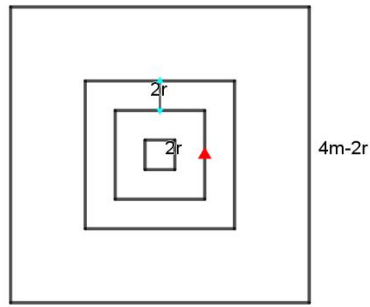

**Fig 3.1.4.2.2** The flying trail of the SSA drones above fire spots

While collecting information, the drones will fly as shown in figure 3.1.4.2.1. The side length of the smallest square is $2r$, and the distance between each concentric square is $2r$. Thus, the recurrence relation is obtained:

$$a_i = 2r + 4(i-1)r$$

(3.1.4.2.2)

Finally, the side length of the last square is:

$$a_m = 2r + 4(m-1)r = (4m-2)\sqrt{d^2 - h^2}$$

(3.1.4.2.3)

Based on the performance of the drone provided by the problem, the total flight time of a drone cannot exceed its maximum endurance time $t$, and the flight speed cannot exceed its maximum flight speed of 72km/h. Therefore, the parameter $m$ can be determined by the following inequality:

$$2 \times m(a_1 + a_m)\sqrt{d^2 - h^2} \leq 72t$$

(3.1.4.2.4)

From this, we can get the mathematical expression of the coverage area of a drone with a task:

$$s_0 = \left(a_m + 2r\right)^2 = 16m^2\left(d^2 - h^2\right)$$

(3.1.4.2.5)

For the safety of frontline firefighters and the complexity of the fire scene, we consider that the interval between collecting data at the same place should not exceed half an hour, in order to continuously collect data, we need a sufficient number of drones as the last batch of used drones. The replacement of the drone when charging.

Finally, we can get the mathematical relationship between the number of drones carrying video and telemetry capability and the area of the forest fire:

$$Q_2 = 2 \times \frac{t}{0.5} \times \frac{S}{s_0} = \frac{tS}{4m^2\left(d^2 - h^2\right)}$$

(3.1.4.2.6)

## 3.1.5 Relationship between the number of drones and fire frequency

After establishing the mathematical relationship between the number of UAVs and the fire size, we discuss the impact of fire frequency on the number of drones. In view of the drone's battery and engine have service life, With the increase of fire frequency, the working frequency of UAVs increases, and the time for batteries and engines to reach their service life shortens, so we need more drones in that case. Base on the data of the wildfire in Victoria Province in Australia last year, we can establish a model of the number of drones and fire frequency.

The average fire-affected area is：

$$\overline{A} = \frac{A}{n}$$

$$(3.1.5.1)$$

The average distance between each fire location and mobile EOC is：

$$\bar{l} = \frac{\sum\limits_{j=1}^{n} l_j}{n}$$

$$(3.1.5.2)$$

Based on the model in 3.3.2, we figure out the average number of drones deployed in response to forest fire：

$$\overline{q_1} = 2\left(\left\lceil \frac{1}{2} \times \frac{\overline{A}}{200} \right\rceil + 1 + \left\lceil \frac{\bar{l}}{20} \right\rceil\right)$$

$$(3.1.5.3)$$

$$\overline{q_2} = \left\lceil \frac{t\overline{A}}{4m^2\left(d^2 - h^2\right)} \right\rceil$$

$$(3.1.5.4)$$

The average time to put out forest fire is：

$$\overline{T} = \frac{T}{n}$$

$$(3.1.5.5)$$

According to the above expression, we can respectively yield the mathematical relationship between the number of drones carrying repeaters and video & telemetry capability and the fire frequency, that is, the number of drones to be purchased in response to the forest fire next year:

$$Q_3 = \overline{q_1}\left\lceil \frac{n\overline{T}}{200} \right\rceil$$

(3.1.5.6)

$$Q_4 = \overline{q_2} \left\lceil \frac{n\overline{T}}{200} \right\rceil$$

(3.1.5.7)

And their ratio is

$$C = \frac{Q_3}{Q_4}$$

(3.1.5.8)

## 3.2 Seasonal Estimated model

### 3.2.1 Model Description

According to the statistics released by the Australian Government, we are told that there is a possibility to have more bushfires in Victoria State. Due to the numbers of bushfires are seasonal, we use the Holt-Winters seasonal model to predict how many fires will happen in Victoria State. This is a cubic exponential smoothing model, it can predict the trend in seasonal function. It contains the predictive function and the cubic exponential smoothing function. The predictive function is indicated in $L_t$, the trending function is indicated in $B_t$, the seasonal function is indicated in $S_t$, also the smoothing variables $\alpha, \beta, \gamma$.

Main function:

$$F_{t+k} = L_t + kb_t + S_{t+k-s}$$

(3.2.1.1)

The meaning of the minor function:

$$L_t = \alpha(y_t - S_{t-s}) + (1-\alpha)(L_{t-1} + B_{t-1})$$

(3.2.1.2)

$$B_t = \beta(L_t - L_{t-1}) + (1-\beta)b_{t-1}$$

(3.2.1.3)

$$S_t = \gamma(y_t - L_t) + (1-\gamma)S_{t-s}$$

(3.2.1.4)

With the constraint:

$$\begin{cases} 0 \le \alpha \le 1 \\ 0 \le \beta \le 1 \\ 0 \le \gamma \le 1 \end{cases}$$

(3.2.1.5)

Furthermore, after we had fit the data with our model, we will use root mean squared error to determine if the data is different from what we expected. The formula can be illustrated in 3.2.5.

$$RMS = \sum_{i=1}^{n}(y_i - y_i^P)^2$$

(3.2.1.6)

## 3.2.2 Notation

| Symbol Name | Definition |
|:---:|:---:|
| $L_t$ | The predictive function |
| $B_t$ | The trending function |

| $S_t$ | The seasonal function |
|---|---|
| $\alpha, \beta, \gamma$ | The smoothing variables |

## 3.3 Drones Deployment Optimization model

### 3.3.1 Image Processing Model

The Image Processing Model is the model that can determine the deployment of drones when fires break out, the size of fires will also affect the result.

First of all, we need to convert the color space of the picture to Hue, Saturation, Value(HSV) color space. The target of this process is to make the picture to be more friendly to our computer. Then we ought to obtain the corresponding range of the color to be extracted by comparing the reference table of HSV. Due to we are extracting the color in both red and yellow, we choose the max hue value of yellow color is 34, the max Saturation value is 255 and the max Value value is 255. Meanwhile, the minimum hue value of the red color is 0, the max Saturation value is 43 and the max Value value is 46.

Afterward, we use the inRange function to extract the color. This function is to extract the color we want and set the color area to white, and the rest to black. To explain the inRange function, we can know that in an RGB three-channel image, this function will let us input a low-value array and a high-value array, and then this function will scan each pixel of the image, the value of each pixel. If the corresponding values are in the values of the corresponding positions in the two arrays we input, then the value will be set to white. Otherwise, as long as one is not in this range, it will be set to black.

### 3.3.2 The best arrangement of drones carrying repeater

In order to make the best arrangement, we figure out the mathematical relationship between the number of drones carrying repeaters for communication and the area of the forest fire:

$$q_1 = \frac{1}{2} \times \frac{S}{200} + 1$$

(3.1.4.1.1)

On the connection line between the fire site and the mobile EOC, the drones with repeaters are arranged in a straight line with a distance of 20 km, which minimizes the cost on the premise of ensuring unblocked communication.

According to the above analysis, we come to the expression of the number of drones on the connecting line between the fire site and the mobile EOC:

$$q_2 = \frac{l}{20}$$

(3.1.4.1.2)

Finally, to keep the temporary communication network unblocked, we need a sufficient amount of drones to replace the last batch of used drones when charging. Therefore, we yield the expression of the number of drones carrying repeaters:

$$Q = 2 \times \left( q_1 + q_2 \right)$$

(3.1.4.1.3)

### 3.3.3 Notation

| Symbol | Definition |
|--------|------------|
| $S$ | The area of fire sense |
| $l$ | The height of the hovering drones |
| $q_1$ | Drones in the fire sense |
| $q_2$ | Drones for connecting EOC and fire sense |

| $Q$ | The total amount of drones |
| --- | --- |

# 4 RESULT AND DISCUSSION

## 4.1 Drones Planning-programming Budgeting Model

After we have analyzed the satellite data from NASA, we can get formula 4.1.1 which contains the value of variables.

$$\begin{cases} A = 600km^2 \\ n = 92 \\ l = 50.5km \end{cases}$$

$$(4.1.1)$$

With the forest fire data, we can know the value of variables is:

$$\begin{cases} h = 0.5km \\ d = 2km \end{cases}$$

$$(4.1.2)$$

After we had applied the data to our model, we can obtain the value of the variables.

$$\begin{cases} r = 1.9365km \\ m = 3 \\ Q_3 = 180 \\ Q_4 = 135 \\ C = 1.33 \end{cases}$$

$$(4.1.3)$$

## 4.2 Seasonal Estimated model

With the aid of python, we can solve the model with just several data set to input, the satellite data is from NASA and we counted the number of fires that happened around Australia. Then we obtain figure 4.2.1 and the Root Mean Squared Error.

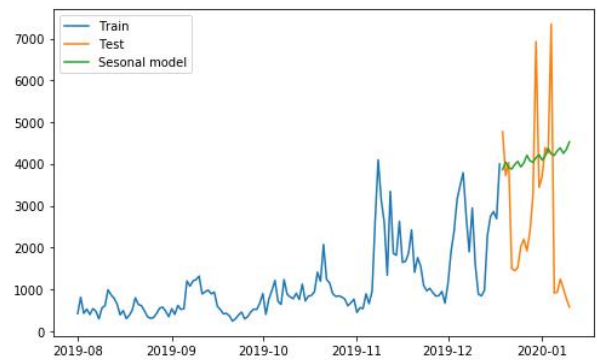

Root Mean Squared = 2325.439068544244

**Fig 4.2.1** The prediction result and the root mean squared error

From figure 4.2.1, we are told that after long-term training for the model, the data of the seasonal model have been proposed, and we can see that it is increasing slowly. Yet, it is similar to sin function for it is periodic. Overall, it shows that the number of fires will be constantly increasing over the next decade. As a result, we assumed that the demand for SSA and repeaters will be constantly growing every year.

Based on the above model, we can infer that the situation of fire may be more serious, the fire-affected area may be larger, and there is a trend of multi-barrier diffusion such as cities and mountains in the next ten years. In addition, we analyze the data of forest fires in the past two years and obtained the trend of forest fires, that is, the frequency of forest fires is likely to increase.

Using the model proposed in section 3.2.1, combined with the above prediction, we can consider the applicability of the model in the next decade.

It is assumed that the cost of the drone system remains unchanged, which includes a drone, a ground-based controller, and a system of communications between the two. The flight of drones may operate with multiple degrees of autonomy: either under remote control by a human operator or autonomously by remote computers referred to as an autopilot[4]. Thus, we do not take anything about drones or their components into account. On the contrary, SSA, repeater, and other equipment are our main consideration.

For SSA, with the increase of fire frequency and area, it is necessary to increase the purchases of SSA to better detect the situation of the fire scene. Moreover, with the increase of SSA's working time and frequency, the faster the battery, engine, and other components reach the expected life, so they need to be purchased more frequently. SSA continues to work under high temperature, it will become more vulnerable, which leads to the increase of SSA repair or maintenance costs. So the government needs to spend more money to make sure SSA can cope with the fire situation in the next decade.

Furthermore, if the scope of the fire is expanded, it may stretch to obstructed areas, such as urban and mountain, which will hinder the transmission of signals of communication equipment. Therefore, we need to buy more repeaters to maintain the transmission of signals. If the number of repeaters is not increased, EOC may not be able to receive the information from the handheld radio in time due to the transmission obstruction area where topography may delay the best rescue time of the fire and lead to greater negative impact. Consequently, it is necessary to increase the budget cost of repeaters.

In addition to the above equipment, other necessary equipment should also be included in the increased budget, such as wearable equipment for firefighters. When the fire frequency increases, the workload of wearable devices increases, and the aging speed is also accelerated. The government needs to spend more money to buy wearable devices, for example, to ensure the construction of a complete emergency fire rescue network.

## 4.3 Drones Deployment Optimization model

## 4.3.1 Image Processing Model

Thanks to OpenCV and NumPy packages, we can solve the problem in a more convenient way. Then, we can process figure 4.3.1.1 to figure 4.3.1.2, which illustrates the range of fires in the Vitoria state.

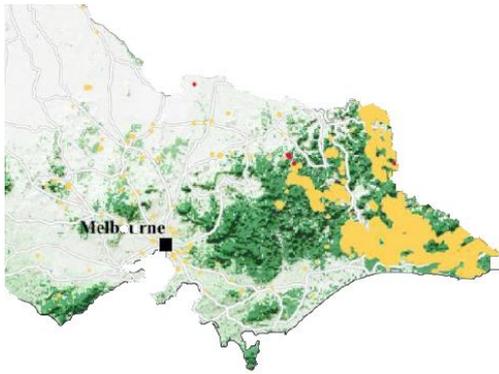 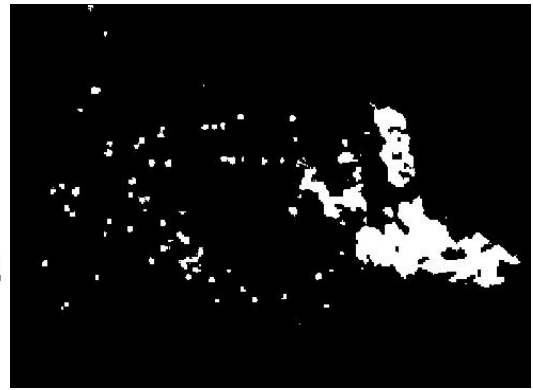

**Fig 4.3.1.2** The original picture        **Fig 4.3.1.2** The processed picture

From figure 4.3.1.2, we can now give the plan of deployment of drones which is to deploy the drones at the edge of the white area.

## 4.3.2 The best arrangement of drones carrying repeater

Assumed that there is a fire event break out as shown in figure 4.3.2.1. We can see that the range of the fire is specified in green color.

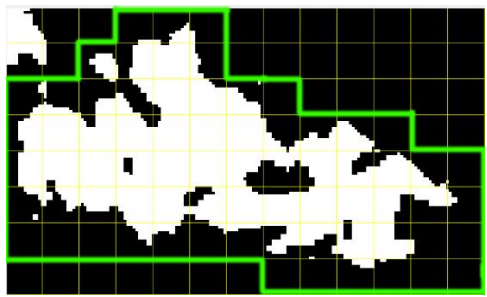

**Fig 4.3.2.1** A hypothetical fire scene

In order to minimize the cost, we arrange drones in the way shown in figure 4.3.2.2 above the fire site on the premise of ensuring unblocked communication.

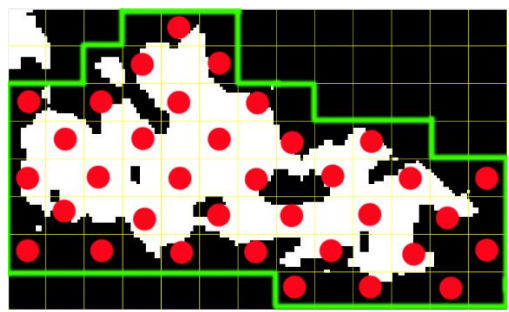

A RAPID WILDFIRE RESPONSE SYSTEM BASED ON DRONES

**Fig 4.3.2.2** The arrangement of drones

(Red points stands for drones)

In another scenario, if firefighters walk into the fire sense for saving people or extinguish fires, their lives might be threatened by the high-temperature of fires. To ensure their safety. We would send drones above them to monitor the status of their equipment and send signals to other drones around. It can be shown in figure 4.3.2.3.

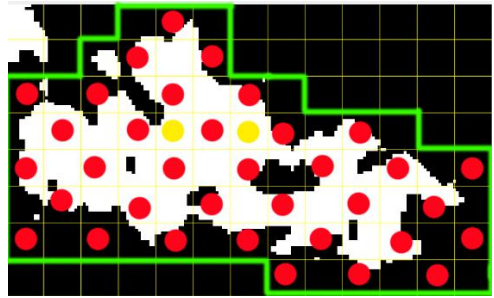

**Fig 4.3.2.3** The arrangement of drones

(Red points represent drones, yellow points represent fireman with drones flying above)

The cells in figure 4.3.2.3 represent a square with a side length of $10\sqrt{2}$ kilometers and an area of 200 square kilometers. The red points represent the hovering drones with the repeaters, and the yellow points represent the mobile drones with the repeaters. For drones arranged in this way, the distance between any two adjacent hovering drones is 20 km, so the signal from a hovering drone can be transmitted to any other hovering drone; the mobile drones ensure that the front-line team's handheld radio signal can be transmitted to hovering drones. Consequently, the drone system ensures unimpeded transmission of signals.

## 5 THE SENSITIVE ANALYSIS

In the sensitivity analysis, we will discuss the sensitivity of the model in four variables: the average fire-affected area $\overline{A}$, the total duration of forest fires $n\overline{T}$, the

A RAPID WILDFIRE RESPONSE SYSTEM BASED ON DRONES

maximum distance of front-line wearable devices on specific terrain $d$, and the height of the hovering drones $h$.

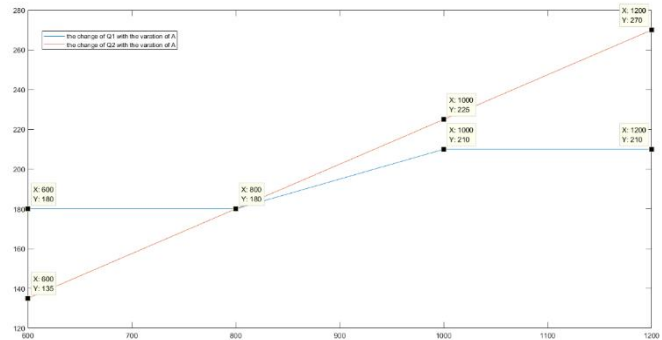

**Figure 5.1** Sensitivity analysis of the average fire-affected area

As shown in figure 5.1, the average fire-affected area has little influence on the number of drones with repeaters. The number remains unchanged when the average fire-affected area increases by one minimum unit of measurement, which indicates that the number of drones with repeaters is not sensitive to the average fire-affected area. Therefore, our model is reasonable. However, the average fire-affected area has a greater impact on the number of drones carrying video and telemetry. When the average fire-affected area increases by a minimum unit of measurement, the drones increases by 45, which indicates that the model still needs to be improved.

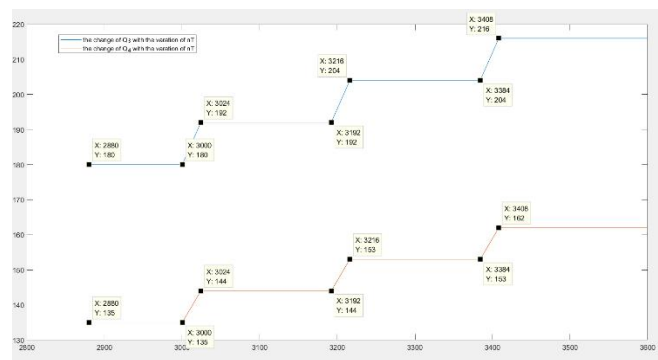

**Figure 5.2** Sensitivity analysis of the total duration of forest fires

As shown in figure 5.2, it turns out that the changes of the two images are relatively gentle. When the change is within 5 days, the number of drones does not change, which indicates that the number of drones is not sensitive to the total duration of forest fires. Therefore, the model is reasonable.

A RAPID WILDFIRE RESPONSE SYSTEM BASED ON DRONES

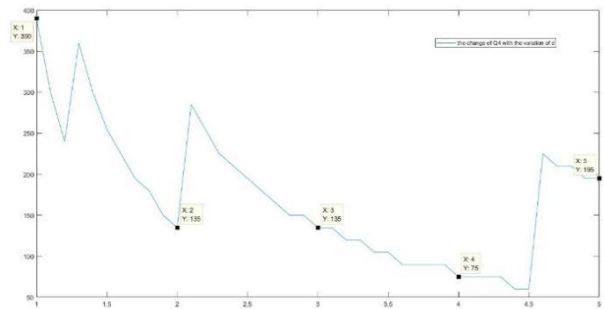

**Figure 5.3** Sensitivity analysis of the maximum distance of front-line wearable devices on specific terrain

As shown in figure 5.3, the change of the image is very steep, which indicates that the number of drones carrying video and telemetry capability is sensitive to the maximum signal transmission distance of front-line wearable devices. As a result, the model needs to consider more influence of geographical conditions.

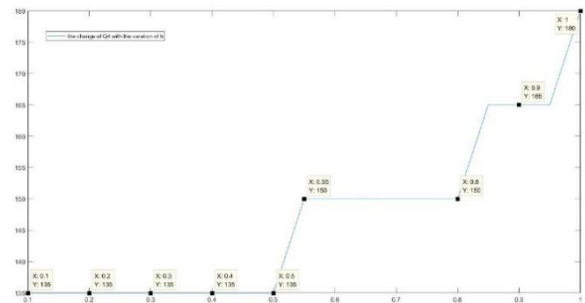

**Figure 5.4** the sensitivity analysis of the height of the hovering drones

As shown in figure 5.4, if the height of the hovering drones is less than or equal to 500m, the number of drones carrying video and telemetry capability remains unchanged, indicating that the number of drones is not sensitive to the height of the hovering drones at low altitude. If the height of the hovering drones is more than 500m, the image changes steeply, which indicates that the number of drones is more sensitive to the hovering height. Consequently,    the model needs to further consider the influence of high altitude.

# 6 STRENGTHS AND WEAKNESS

## 6.1 Strengths

Our model is robust with all inputs based on real-world data from trusted websites and governmental agencies such as the Country Fire Authority and South Australian Government Data Directory. Besides, our model contrapuntally focused on the Victoria State, we have referred a great amount of data about bushfires in Victoria. With a variety of data set of information that we accumulate, we can use our model to estimate the quantity of unmanned aerial vehicles. Moreover, we consider the geographical environment of the fire location, fire-affected area, and frequency, which is closer to the reality than the model considering only a single factor. In addition, we also take into account the fire situation in Victoria State in the past year, which makes our model more reasonable. Last but not least, we give a fast response mechanism for the forest fire. The rescue team can firstly carry the equipment to the fire sense. Afterward, the drones will transmit at the mobile EOC, with the signal receiving terminal where the command of front-line members is completed. As a result, we are able to provide a reference method for the emergency response mechanism of the Victoria Country Fire Authority.

As for the seasonal Estimated model, we have mentioned the seasonal estimation model, this model is based on the Holt-Winters' seasonal method and being developed by us. It has used a great amount of historical bushfires data, so we can analyze them and use them to train our model. Then, fit the data with our trained model. So we consider that it is respectively correct to other mathematical models for machine learning are in use. In addition to it, we consider our model takes various situations into account for the exponential smoothing function can use in different situations. Yet, it is easy to understand for the formulation of this model is simple. Plus, it is user-friendly, we just need to determine the data that we input so that we can use it freely.

Next, the Color Space Estimated model, we consider its strength is it is simple to understand and it is convenient to use it for we just need to use the packages in Python so that we can estimate the area aims to help us to deploy the unmanned aerial vehicles.

## 6.2 Weaknesses

Due to the limited data available, we can not give accurate results because of the large error in solving the model. Besides, the mathematical model of the influence of geographical environment on the range of front-line wearable devices and handheld radios is relatively simple, which leads to a certain deviation from the actual situation. Moreover, in the model that evaluate the numbers of drones carrying video and telemetry capability, we did not further discuss and analyze the hovering height $h$ and mission given time $t$ of drones, which is not beneficial to provide more reference for the establishment of rapid response mechanism of forest fire in the new circumstance.

Regarding the Seasonal Estimated Model, we consider that the weaknesses of this model are the variables that need to take a long time to be verified by a variety of experiments, we just have few days to finish the model so we can not make sure it is 100% correct. And we use the fixed data to make the combination of the minimum loss, which means the result can be local but not global. In other words, it leads to the data can be overfitting which makes the result is not as expected.

Regarding the Color Space Estimated model, we consider that the weakness of this model is it does not take too many things into account such as the height. Without the involution height, we may deploy the drones in an inefficient way which means we would spend more money on it.

# 7 CONCLUSION

In the 2019-2020 fire season in Australia, we can see a variety of wildfires happened in every state, it can cause severe property damage, the lives of wildlife and human are being threatened as well. Thus, to fight wildfires, our team is asked to solve three problems, the first one is to estimate the quantity of SSA drones and repeater drones that should be purchased by the CFA. With the selections of mobile EOC sites accomplished, We propose the Drones Planning-programming Budgeting Model to estimate the cost and quantities of each drone. Furthermore, we use the model to give out a "Rapid Bushfire Response" to the CFA. Fire event size and frequency are also being considered in our first model.

In order to solve the second problem which is how our model can adapt to the extreme situation in the next ten decades, we build a Seasonal Estimated Model to forecast the future trend of fire, including its frequency and spreading area. Based on the prediction of this model, we project what equipment cost increases will occur assuming the cost of drone systems stays constant.

For question3, We adapt the Image Processing Model model to locate the fire sites clearly in the picture given. Combined with the expression in our first model, we can optimize the best arrangement of drones carrying repeater by the Drones Deployment Optimization model. Thus, we make a detailed strategic deployment for the hovering position of the drones.

According to the three models discussed above, we can construct a "Rapid Bushfire Response" system. The mobile EOCs, responsible for the launching of the drone, is located in Mildura, Horsham, Portland, Ballarat, Melbourne, Wangaratta, Wodonga, Sale, and Bairnsdale. Every mobile EOC can deploy the drones with a total of 315, including 185 drones carrying repeaters and 130 drones carrying telemetry capability. As for the hovering position planning of drones, we make the distance between any two adjacent hovering drones be 20 km, so the signal sent by one hovering drone can be transmitted to any other hovering drone. On the other hand, the mobile drones ensure that the handheld radio signal of the front-line personnel can be transmitted to the hovering drones. Therefore, the drone's communication system ensures unimpeded signal transmission. The adaptability improvement is also involved in the system. We can make it apply to the extreme circumstance in the future by increasing the number of SSA and repeaters, because the fire-affected area may be larger, the fires happen more often, and there is a trend of multi-barrier diffusion such as cities and mountains in the next ten years. Therefore, we can establish an effective fire emergency treatment method within ten years to ensure that the fire can be detected, monitored, and solved in a very short time to minimize its adverse influence on Victoria State.

Our follow-up work is to continue to optimize the model. In the process of drone deployment, the terrain is an important factor. In the following work, we need to evaluate the impact of topography on drone position and modify the design scheme of drone position based on it, so as to make the drone's layout more suitable for the real situation and deal with the practical problems reasonably.

To conclude, this fire emergency response mechanism enables us to cope with the fire emergency expectedly, which provides the Australia government with a feasible reference to the response to the fire, and contributes a theoretical math modeling scheme in the real world.

A RAPID WILDFIRE RESPONSE SYSTEM BASED ON DRONES